\documentclass[letterpaper]{article} 
\usepackage{aaai2026}  
\usepackage{times}  
\usepackage{helvet}  
\usepackage{courier}  
\usepackage[hyphens]{url}  
\usepackage{graphicx} 
\usepackage{natbib}  
\usepackage{caption} 
\usepackage{algorithm}
\usepackage{algorithmic}
\usepackage{amsmath,amssymb,amsfonts}
\usepackage{placeins}
\usepackage{graphicx}
\usepackage{textcomp}
\usepackage{booktabs} 
\usepackage{xcolor}
\def\BibTeX{{\rm B\kern-.05em{\sc i\kern-.025em b}\kern-.08em
    T\kern-.1667em\lower.7ex\hbox{E}\kern-.125emX}}
\newcommand{\ph}[1]{\texttt{\$\{\detokenize{#1}\}}}
\usepackage[most]{tcolorbox}
\newtcolorbox{promptbox}[1][]{
  colback=gray!10,
  colframe=gray!80,
  boxrule=0.5pt,
  arc=4pt,
  left=4pt,      
  right=4pt,     
  top=3pt,       
  bottom=3pt,    
  boxsep=4pt,    
  breakable,
  before skip=6pt,    
  after skip=6pt,     
  enhanced,
  colbacktitle=gray!15,
  coltitle=black,
  fonttitle=\bfseries\color{black},
  title={#1}
}

\usepackage{newfloat}
\usepackage{listings}
\DeclareCaptionStyle{ruled}{labelfont=normalfont,labelsep=colon,strut=off} 
\floatstyle{ruled}
\newfloat{listing}{tb}{lst}{}
\floatname{listing}{Listing}
\title{Adaptive Selection of Symbolic Languages for Improving LLM Logical Reasoning}
\author{
Xiangyu Wang\textsuperscript{\rm 1}, Haocheng Yang\textsuperscript{\rm 2}, Fengxiang Cheng\textsuperscript{\rm 3}\thanks{Fengxiang Cheng and Fenrong Liu are corresponding authors.}, Fenrong Liu\textsuperscript{\rm 1,3}$^\ast$}
\affiliations{
    \textsuperscript{\rm 1}Tsinghua University
    \textsuperscript{\rm 2}National University of Singapore
      \textsuperscript{\rm 3}University of Amsterdam\\
      f.cheng@uva.nl, fenrong@tsinghua.edu.cn
}

\usepackage{bibentry}

\usepackage{csquotes}

\begin{document}
\maketitle
\begin{abstract}
Large Language Models (LLMs) still struggle with complex logical reasoning. While previous works achieve remarkable improvements, their performance is highly dependent on the correctness of translating natural language (NL) problems into a symbolic language (SL). Though numerous works focusing on improving this translation accuracy, they only consider the similarity between the meaning of SL and NL, overlooking another crucial influencing factor, the selection of the target SL type itself. For example, first-order logic language specializes in logical reasoning with categorical syllogisms and complex quantifiers, while Boolean satisfiability formalism excels at representing constraint satisfaction like partial problems.
To our knowledge, this is the first paper to claim and verify that different NL logical reasoning problem corresponds to different optimal SL formalization for translation. Based on this, we propose a methods to improve the logical reasoning performance of LLMs by adaptively selecting the most suitable SL for each problem prior to translation. Specifically, we leverage LLMs to select the target SL among first-order logic, logic programming and Boolean satisfiability and then translate the problem in NL to target SL expressions as well as employ the corresponding logical solver to derive the final answer. Experimental results on benchmarks show that our adaptive selection method significantly outperforms translating all into single SL and randomly selecting the SL. On a mixed dataset of these benchmarks, our approach achieves 96.00\% accuracy, which improving performance by 25\% compared to the second highest accuracy from the first-order logic translation.
\end{abstract}
\section{Introduction}

Large language models (LLMs) have exhibited outstanding performance on diverse natural language processing tasks, but they still face significant challenges in complex logical reasoning, which limits their practical applicability in real-world scenarios~\cite{cheng2025empowering,cheng2025mitigating,lv2025breaking,yu2025cauosal}. Efforts to enhance the logical question answering (QA) abilities of LLMs can be categorized into three  approaches: prompt-based methods~\cite{xu2024faithful, xu2024aristotle}, fine-tuning methods~\cite{morishita2024enhancing,wan2024logicasker}, and external solver-based methods~\cite{ye2023satlm, ryu2024divide}.
Prompt-based methods leverages LLMs to translate and reason logical question-answering problems directly while fine-tuning approaches enhance logical reasoning performance by constructing synthetic datasets that expose detailed logical deduction steps. This work focuses on the solver-based methods, which transform natural language (NL) questions into symbolic language (SL) expressions before employing specialized solvers for inference.

Employing an external solver offers reliable reasoning, as these solver tools provide deterministic and verifiable execution~\cite{olausson-etal-2023-linc,ye2023satlm}. 
However, reasoning performance of logical solvers is acutely sensitive to the accuracy of the initial translation, often resulting in low execution rates and overall accuracy rate due to parsing failures. To enhance translation quality, numerous approaches have been proposed, including translating the raw NL paragraph to simpler and atomic NL subsentences~\cite{ryu2025divide}, self-refinement loops where feedback from a logical solver is used to correct erroneous logical statements~\cite{callewaert2025verus}, and bidirectional translation checks (SL→NL→SL) to automatically verify logical equivalence without human annotation~\cite{karia2024forall}. 



However, previous works only focus on one factor affecting the accuracy of translation results, i.e., the similarity between expressions in this fixed SL and the NL representation of the target question, while overlook another crucial influencing factor, namely selecting the type of SL that is most suitable to formalize the target question. 
For instance, First-Order Logic (FOL) language excels at addressing logical reasoning problems required categorical syllogisms with complex quantifiers, whereas the Boolean Satisfiability (SAT) formalism is highly proficient in representing constraint satisfaction tasks such as ordering problems.
Even though when semantic similarity is high, if adopting an inappropriate SL type, which is unable to properly formalize and capture all content of the NL problem, it will still lead to both lower translation and reasoning accuracy.
We found that for the LogicalDeduction dataset, translating all questions into SAT yields 90\% reasoning accuracy with the corresponding solver, while FOL translation only gives 42\%; conversely, for ProofWriter, FOL achieves 95.50\% accuracy but SAT drops to 68.33\%.
Further details will be provided in the experiments.
We are the first paper to claim that the selection of the symbolic language significantly impacts translation accuracy and consequently the reasoning performance including the execution rate of solvers. 
This intuitively demonstrates that different formalisms—such as FOL, SAT or Logic Programming (LP)—are suited to different logical QA problems due to the different expressive capabilities and complexity of these languages.

Therefore, to fill this research gap, we then
propose an approach that adaptively selects the most appropriate SL for a given reasoning problem. Our method prompts an LLM to choose the most appropriate SL for each problem by comparing their expressive features, advantages, and disadvantages of several candidate SL. Subsequently, the model then translates the NL query into the chosen SL, and a corresponding logic solver is employed to derive the final answer. 
Through experiments, we demonstrate the remarkable effectiveness and feasibility of our proposed method.


Our main contributions can be summarized as follows:

\begin{itemize}
    \item To the best of our knowledge, we are the first to claim and verify experimentally that each NL logical reasoning problem corresponds to an optimal SL for translation.
    \item We propose an adaptive selection method that significantly enhances NL-to-SL translation accuracy, the execution rate of solvers, and the correctness of the final reasoning outcomes. 
    \item We design experiments showing that (1) comparisons across three types of SL translations of the same dataset confirm that each logical problem has an optimal SL, and (2) our methods of adaptive selection of SL significantly outperforms random choice.
    
\end{itemize}


\section{Related Work}
\subsection{Logical Reasoning in LLMs}
Efforts to enhance the logical reasoning capabilities of LLMs can be categorized and summarized into three approaches: solver-based, prompt-based, and fine-tuning methods~\cite{cheng2025empowering,cheng2025enhancing}. 
Solver-based methods initially translate NL queries into SL formulations and subsequently leverage dedicated solvers to execute the inference task \cite{lyu2023faithful, ye2023satlm, olausson-etal-2023-linc, ryu2024divide}. Prompt-based techniques follow two main pipelines: one involves the explicit generation of NL reasoning steps to derive the conclusion \cite{wei2022chain,yao2024tree,zhang2024diagram}, while the other prompts LLMs to perform NL-to-SL translation, sequential reasoning, and answer validation \cite{xu-etal-2024-symbol,liu2024logicofthoughtinjectinglogiccontexts,li2024leveraging,fu2025improve}. Finally, fine-tuning strategies enhance LLMs' logical reasoning capability either by synthetically creating datasets that feature logical derivation proofs \cite{bao-etal-2024-abstract,morishitaenhancing} or by augmenting existing training corpora with logical reasoning examples \cite{feng-2024-languagecanbe,wan-etal-2024-logicasker,jiao2024exploring}.
In contrast of these methods generally considered only one SL formalization for a logical QA, our work introduces a crucial preliminary step before the translation process: adaptively selecting the most suitable SL for each distinct problem.


\subsection{Methods to Improve the Accuracy of Translation}
A primary challenge for solver-based reasoning systems is the accuracy of the translation from NL to SL \cite{lyu2023faithful, ye2023satlm, olausson-etal-2023-linc, ryu2024divide}. The inference performance of solvers is critically dependent on the correctness of this initial translation step. Several strategies have been proposed to mitigate translation errors. CLOVER~\cite{ryu2025divide} decomposes complex NL statements into simpler units before converting them into the SL. VERUS-LM~\cite{callewaert2025verus} employs an iterative refinement loop where the reasoning engine provides feedback to correct both syntactic and semantic errors in the generated logical forms. $\forall$uto$\exists$val~\cite{karia2024forall}, proposes a self-verification technique based on a round-trip translation (SL→NL→SL) to check for logical consistency without manual supervision. While these techniques focus on refining the translation process, our approach addresses a more fundamental aspect: selecting the optimal target SL before translation begins, which directly influences both translation feasibility and accuracy.

\section{Logical Question Answering Problem Setup}
Logical QA tasks are centered around assessing that: according logical inferrence rules, whether a given statement can be validly inferred from provided contextual information through strict logical deduction. For LLMs tackling such tasks, the core requirement is to accurately answer the truth status of this statement among three options: \textit{true} (when the statement can be logically deduced by the premises), \textit{false} (when it contradicts the given information), or \textit{unknown} (when insufficient evidence exists to confirm or refute it). An illustrative example is provided below.

\begin{promptbox}[An example for Logical QA]
{\bf Contexts (Premises):}

$\bullet$ The tiger is big.\\
$\bullet$ If something is big then it visits the rabbit.\\
$\bullet$ The rabbit visits the tiger.\\
$\bullet$ If something visits the rabbit then the rabbit needs the lion.\\
$\bullet$  If something sees the tiger then it is rough.\\
$\bullet$ ......\\
{\bf Question:} Based on the above information, is the following statement true, false, or unknown? The rabbit does not need the lion.

{\bf Options:} A) True \quad B) False \quad C) Unknown

{\bf Answer:} B) False
\end{promptbox}




\begin{figure*}
    \centering
\includegraphics[width=1\linewidth]{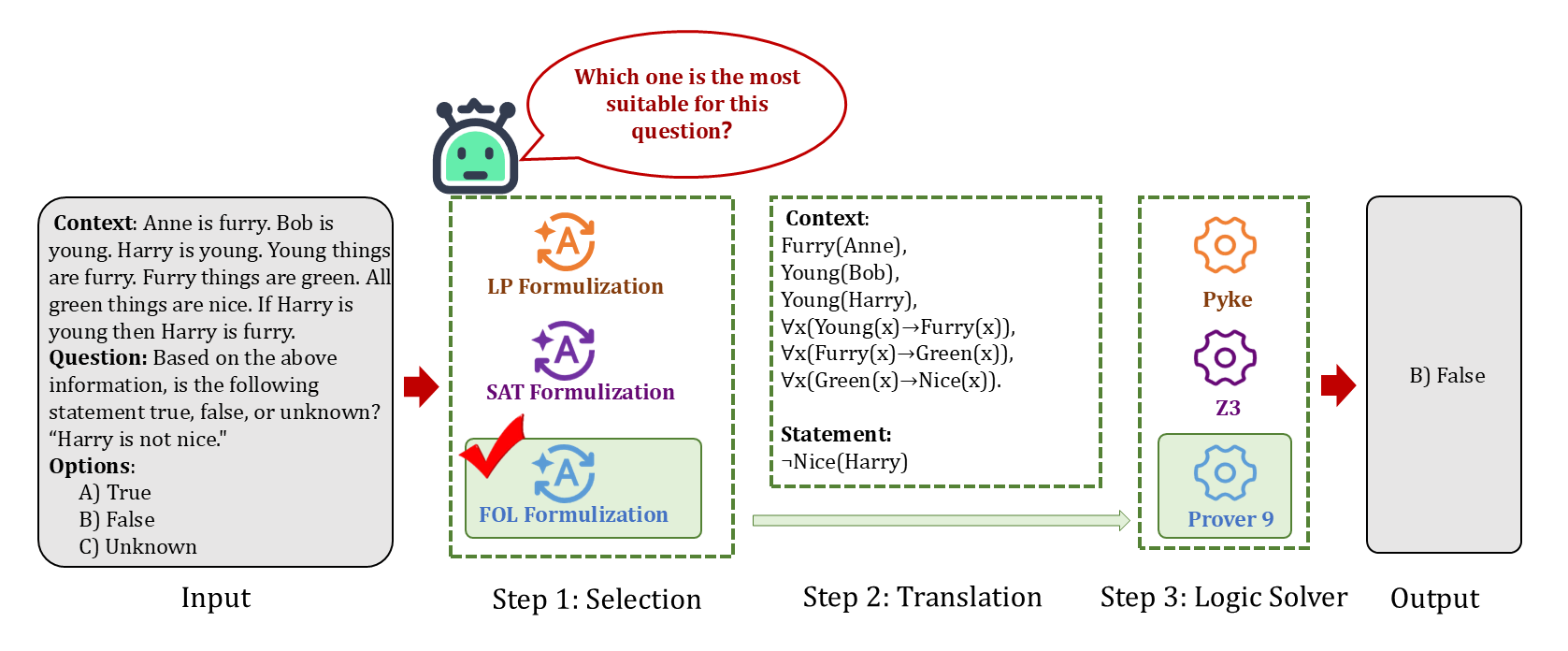}
\vspace{-20pt}
    \caption{The framework of our methods to adaptively select symbolic languages to translate logical reasoning problems}
    \label{fig:framework}

\end{figure*}

\section{Proposed Method}
Our method for solving logical QA tasks consists of three main stages: an adaptive symbolic language selection step, a translation step, and a reasoning step. As illustrated in Figure~\ref{fig:framework}, given a natural language logical reasoning problem, we prompt an LLM first determines the most suitable SL among three candidates: FOL, LP, and SAT. The problem is then translated into the chosen SL by the LLM. Finally, an external logical solver is used to perform the reasoning and generate the final answer. This pipeline ensures that each problem is processed using the most appropriate formalization, leading to higher accuracy of final answer.

\subsection{Three Formalization of Symbolic Languages}
We use three distinct symbolic languages tailored for different types of logical reasoning problems. \textbf{FOL} is suitable for problems involving complex quantification and relationships between entities. Its expressive power allows for a direct translation of premises and rules that contain universal or existential quantifiers. \textbf{LP} are well-suited for deductive reasoning based on a set of facts and rules. They are particularly effective for problems where a clear chain of forward or backward inference can be established. \textbf{SAT} formalizes problems as constraints to check for satisfiability. It is highly efficient for problems that can be reduced to boolean constraints, such as those with a large number of relationships.

\begin{table*}[th!]
\centering
\caption{Performance comparison of different symbolic language selection strategies on three benchmarks. The best results are highlighted in bold.} 
\label{tab:model_performance} 
\resizebox{\textwidth}{!}{%
\begin{tabular}{l|ccc|ccc|ccc}
\toprule
& \multicolumn{3}{c|}{\textbf{ProntoQA}} & \multicolumn{3}{c|}{\textbf{ProofWriter}} & \multicolumn{3}{c}{\textbf{LogicalDeduction}} \\
\cmidrule(lr){2-4} \cmidrule(lr){5-7} \cmidrule(lr){8-10}
& Overall-Acc & Exec-Rate & Exec-Acc & Overall-Acc & Exec-Rate & Exec-Acc & Overall-Acc & Exec-Rate & Exec-Acc \\
\midrule
Chance & 50.00\%  & / & / & 33.33\% & / & / & 20.00\% & / & / \\
LP & 93.60\%  & 87.80\%  & 99.66\%  & 78.83\%  & 98.50\%  & 79.53\%  & 36.33\%  & 74.33\%  & 41.97\%  \\
FOL  & 98.80\%  & 98.60\%  & 99.59\%  & 95.50\% & 97.83\% & 96.88\%  & 42.00\%  & 32.00\%  & 88.75\%  \\
SAT & 80.60\%  & 65.20\%  & 96.93\%  & 63.83\% & 68.33\%  & 77.97\%  & 90.00\%  & 93.67\%  & 94.73\%  \\
Random selection & 88.80\%  & 82.20\%  & 97.20\%  & 77.50\%  & 86.83\%  & 84.20\%  & 53.33\%  & 63.67\%  & 72.36\% \\
Adaptive selection & \textbf{99.80\%}  & \textbf{98.80\%} & \textbf{100.00\%} & \textbf{96.00\%}  & \textbf{98.17\%}  & \textbf{97.17\%}  & \textbf{91.33\%} & \textbf{94.33\%} & \textbf{95.62\%} \\
\bottomrule
\end{tabular}%
}
\vspace{-5pt}
\end{table*}

\begin{table}[th!]
\centering
\caption{Performance comparison on the mixed dataset} 
\label{tab:mixed_data_performance} 
\begin{tabular}{l|ccc}
\toprule
& \multicolumn{3}{c}{\begin{tabular}[c]{@{}c@{}}\textbf{Mixed} \end{tabular}} \\
\cmidrule(lr){2-4}
& Overall-Acc & Exec-Rate & Exec-Acc \\
\midrule
LP & 69.33\%  & 83.33\%  & 76.32\% \\
FOL & 79.67\%  & 76.00\%  & 93.96\% \\
SAT & 78.67\%  & 77.00\%  & 91.89\%  \\
Random selection & 70.67\%  & 74.00\%  & 83.41\% \\
Adaptive selection & \textbf{96.00\%} & \textbf{96.33\%} & \textbf{98.34\%} \\
\bottomrule
\end{tabular}
\vspace{-5pt}
\end{table}

\subsection{Adaptively Selection to the Symbolic Languages}
Our framework employs an adaptive selection mechanism that prompts a LLM to choose the most suitable SL for each problem. The prompt provides the LLM with a comparative analysis of the candidate SLs, detailing their distinct expressive features, advantages, and disadvantages. Based on this information and the specific problem at hand, the LLM determines which language—FOL, LP or SAT—is best suited for the translation and reasoning task.

\begin{promptbox}[ Prompts for Adaptively Selection of SL]
You are an expert in symbolic logic and reasoning systems. Your task is to analyze a logic problem and select the most appropriate symbolic language for solving it.

You have three symbolic languages to choose from:
\begin{enumerate}
    \item FOL (First-Order Logic):\\
\textbf{-Best for}: Complex quantifiers, mathematical relationships, formal proofs.
\textbf{-Features}: Universal ($\forall$) and existential ($\exists$) quantifiers, logical operators ($\lnot,\vee, \wedge, \rightarrow$), predicates, functions, variables.
\textbf{-Typical problems}: Mathematical theorems, complex logical relationships, nested quantifications, categorical syllogisms.
\textbf{-Example patterns}: ``For all $X$, there exists $Y$ such that...", ``If and only if...", ``All $X$ are $Y$".

\item \textbf{LP (Logic Programming)}: \\
\textbf{-Best for}:  Deductive reasoning, propositions, relationship between sentences.
\textbf{-Features}: Fact as a simple statement with  predicates and arguments. Rules written in the form of clauses. Query as another fact required to be proved based on known facts and rules.
\textbf{-Typical problems}: Deductive reasoning, propositional logical reasoning.
\textbf{-Example patterns}: ``If something is $X$ then it is $Y$".

\item \textbf{SAT ( Boolean Satisfiability Problem)}:
\textbf{-Best for}: Constraint satisfaction, spatial/ordering problems, discrete choices.
\textbf{-Features}: Boolean variables, constraints, position/ordering relationships.
\textbf{-Typical problems}: Arrangement puzzles, scheduling, spatial reasoning.
\textbf{-Example pattern}s: ``$X$ is to the left of $Y$", "$X$ is between $Y$ and $Z$".

\end{enumerate}

Given the following logic problem:

Context: \ph{context}\\
Question: \ph{question}\\
Options: \ph{options}\\

Analyze the problem structure carefully and select the symbolic language that best matches the problem.


\end{promptbox}


\subsection{Translation via LLMs and Reasoning via Solvers}
The translation from natural language to the chosen symbolic language is performed by a LLM. This LLM is prompted to convert the given premises and question into a syntactically correct and semantically faithful expressions in the selected SL. The generated symbolic expressions are then passed to their corresponding external solvers. Specifically, we use \textbf{Prover9}~\cite{prover9} for FOL, \textbf{Pyke}~\cite{pyke} for LP, and \textbf{Z3}~\cite{de2008z3} for SAT. These solvers execute the logical reasoning, such as theorem proving or satisfiability checking, and return a definitive logical result. This result is then transformed back to the final answer (e.g., True, False, or Unknown).

\section{Experiments}
\subsection{Experimental Setup}
The experiments are conducted on GPT-4~\cite{openai2023gpt4}. We evaluate our method on three distinct logical reasoning benchmarks: ProntoQA~\cite{saparovlanguage}, ProofWriter~\cite{tafjord2020proofwriter}, and LogicalDeduction~\cite{srivastava2023beyond}. The performance is measured by three metrics: Overall-Acc (Overall Accuracy), Exec-Rate (Execution Rate, representing the proportion of total samples for which the solver can read and execute reasoning on the translated results), and Exec-Acc (Execution Accuracy, estimating the proportion of samples with correct reasoning results relative to the samples can be reasoned by the solver). We compare our adaptive selection method against four baselines, including translating all samples into a single SL (including LP, FOL and SAT) and a random selection baseline. A further test is performed on a Mixed dataset constructed by selecting 100 samples in each dataset (ProntoQA, ProofWriter, and LogicalDeduction) to evaluate the LLM' performance in a more generalized setting.

\subsection{Results Analysis}
The results from Table \ref{tab:model_performance} demonstrate that our adaptive selection methods performs best on all benchmarks among translating all samples into one single SL and randomly selecting the SL. It is also verified our claim that different NL logical reasoning problem corresponds to different optimal SL formalization for translation. 
Apart from our adaptive selection methods, we find that on the three datasets of ProntoQA, ProofWriter, and LogicalDeduction, FOL (corresponding to the first two) and SAT respectively achieve the highest accuracy, with Overall-Acc of 98.80\%, 95.50\%, and 90.00\% respectively.
This highlights that
a ``one-SL-fits-all" approach to translate is sub-optimal, and 
selecting the appropriate language is crucial for maximizing performance on different logical QA problems.

The comparison between the random selection and adaptive selection methods confirms the effectiveness of our approach. As shown in both Table \ref{tab:model_performance} and Table \ref{tab:mixed_data_performance}, the adaptive selection method consistently outperforms the random selection method across all benchmarks and the mixed dataset. On the Mixed dataset, adaptive selection achieves a remarkable 96.00\% Overall-Acc, a significant improvement over the 70.67\% for Random selection. This dramatic difference demonstrates that intelligently selecting the most suitable symbolic language for each problem instance is far superior to approaches which doesn't consider the SL selection and validates the effectiveness of our adaptive selection method.

\section{Conclusion}

In this paper, we are the first to claim that the optimal choice of SL for solver-based logical reasoning dependent on each problem is crucial to the performance of both translation and final reasoning. We introduce an approach to improve LLMs' logical reasoning capabilities by prompting LLM adaptively selects the most suitable formalism—from FOL, LP, SAT—for each unique problem before the translation process. Our experiments confirm the effectiveness of our approach, showing that it significantly improves overall accuracy compared to fixed-SL and random-selection baselines. This work establishes that treating SL choice as a dynamic decision is a vital step to improve LLMs logical reasoning abilities. For future work, we plan to expand our framework by exploring a broader range of SL and aim to establish a more formal, theoretical foundation to guide the SL selection process.

\bibliography{aaai2026}

\end{document}